\documentclass[journal]{IEEEtran}

\usepackage{graphicx}
\usepackage{amsmath}
\usepackage{amssymb}
\usepackage{fontawesome}
\usepackage{algorithm}
\usepackage{algorithmic}
\usepackage{tabularx,booktabs}
\usepackage{enumitem}
\usepackage{multirow}

\usepackage{soul}
\usepackage[breaklinks=true,colorlinks,bookmarks=false]{hyperref}
\usepackage{cite}

\usepackage[table]{xcolor}
\definecolor{shadecolor}{rgb}{0.95,0.95,0.92}
\definecolor{codegreen}{rgb}{0,0.6,0}
\definecolor{codegray}{rgb}{0.5,0.5,0.5}
\definecolor{codepurple}{rgb}{0.58,0,0.82}
\definecolor{backcolour}{rgb}{0.95,0.95,0.92}
\definecolor{dukeblue}{rgb}{0.0, 0.0, 0.61}
\definecolor{midnightblue}{rgb}{0.1, 0.1, 0.44}
\definecolor{navyblue}{rgb}{0.0, 0.0, 0.5}
\newcommand{\coloururl}[1]{{\color{magenta} #1}}

\definecolor{greenPython}{RGB}{0,150,0}

\def\qut#1{\left(#1\right)}

\def\qutl#1{\mathopen{}\left(#1\right)\mathclose{}}
\def\qutc#1{\left\{#1\right\}}
\def\matL{\mathcal{L}}

\newcommand{\blockb}[3]{\multirow{3}{*}{\(\left[\begin{array}{c}\text{1$\times$1, #2}\\[-.1em] \text{3$\times$3, #2}\\[-.1em] \text{1$\times$1, #1}\end{array}\right]\)$\times$#3}
}

\ifCLASSINFOpdf
\else
\fi

\begin{document}
\title{A Three-Stage Self-Training Framework for Semi-Supervised Semantic Segmentation}

\author{
        Rihuan Ke\textsuperscript{*}, 
        Angelica I. Aviles-Rivero\textsuperscript{*},
        Saurabh Pandey,
        Saikumar Reddy,
        Carola-Bibiane Sch{\"o}nlieb%
\thanks{
R. Ke, A. Aviles-Rivero and C.-B. Sch{\"o}nlieb are with  Centre for Mathematical Sciences, University of Cambridge, Cambridge CB3 0WA, U.K.}%
\thanks{S. Pandey and S. Reddy are with Kritikal Solutions Pvt. Ltd., Bangalore, India.}%
}

\markboth{}%
{Shell \MakeLowercase{\textit{et al.}}: }

\maketitle
\begingroup\renewcommand\thefootnote{*}
\footnotetext{Equal contribution}
\endgroup

\begin{abstract}
Semantic segmentation has been widely investigated in the community, in which state-of-the-art techniques are based on supervised models. Those models have reported unprecedented performance at the cost of requiring a large set of high quality segmentation masks for training.  Obtaining such annotations is highly expensive and time consuming, in particular, in semantic segmentation where pixel-level annotations are required. In this work, we address this problem by proposing a holistic solution framed as a 
self-training framework for semi-supervised semantic segmentation. 
The key idea of our technique is the extraction of the pseudo-mask information on unlabelled data whilst enforcing segmentation consistency in a multi-task fashion. 
We achieve this through a three-stage solution. 
Firstly, a segmentation network is trained using the labelled data only and rough pseudo-masks are generated for all images. 
Secondly, we decrease the uncertainty of the pseudo-mask by using a multi-task model that enforces consistency and that exploits the rich statistical information of the data. 
Finally, the segmentation model is trained by taking into account the information of the higher quality pseudo-masks. 
We compare our approach  against existing semi-supervised semantic segmentation methods and demonstrate state-of-the-art performance with extensive experiments. 
\end{abstract}


\IEEEpeerreviewmaketitle

\section{Introduction}
\IEEEPARstart{S}{emantic} segmentation is a fundamental task in image analysis. It aims to assign a label, from a set of predefined classes, to each pixel in the image. This task has been widely explored in the literature yet not fully solved.  The current state-of-the-art models are based on building upon deep nets~\cite{long2015fully,chen2017deeplab,noh2015learning,zhao2018psanet,Zhen_2020_CVPR}. Whilst these techniques have reported unprecedented results, they rely on a very high labelled data regime. This is a strong assumption as annotations are pixel-level, which are expensive, time consuming and inherent to human bias.  To address the lack of a large and well-representative set of labels, one could rely more on unlabelled samples.

An alternative is to use solely  unlabelled data -- i.e. unsupervised learning. However, this paradigm has not been successful for semantic segmentation as the performance substantially degrades due to  the lack of correspondence between the samples and classes.
Another option is to use weakly supervised techniques~\cite{lee2019ficklenet,shimoda2019self}, nonetheless, the rich information from unlabelled samples is not fully exploited and performance is still limited.
A feasible option is to use semi-supervised learning that leverages on a vast amount of labelled data and a tiny set of annotations. Although  semi-supervised learning (SSL) has been widely developed~\cite{chapelle2009semi} in the community, the progress of deep semi-supervised learning has been only noticeable in the last few years, and mainly for the task of image classification e.g.~\cite{laine2016temporal,miyato2018virtual,verma2019interpolation}, 
the principle of which have been used recently in the context of semantic segmentation e.g.~\cite{hung2019adversarial,french2019semi,ouali2020semi,feng2020semi}.

The existing techniques for SSL can be broadly divided in entropy minimisation~\cite{grandvalet2005semi}, generative models~\cite{kingma2014semi}, graph based techniques~\cite{zhou2003learning}, proxy-based techniques (building upon pseudo-labels/self-training)~\cite{xie2020self}, consistency regularisation~\cite{laine2016temporal,miyato2018virtual} and holistic approaches that take the best of each principle. For semantic segmentation existing techniques use generative models and consistency regularisation techniques. Although the results reported for this task are promising, there is plenty of room for improvement, and in particular, regarding how to improve the confidence score predictions.

\begin{figure}
    \centering
    \includegraphics[width=\linewidth, trim=160 150 230 120, clip]{{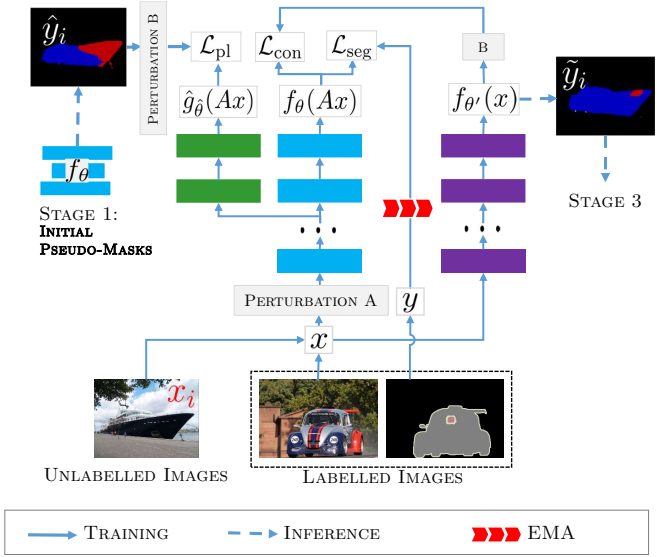}}
    \caption{
    The proposed self-training framework (the first two stages are drawn). Stage 1 uses a segmentation network $f_\theta$ trained on the tiny labelled set to produce initial pseudo-masks. For Stage 2, the pseudo-masks are used to learn an auxiliary task (with loss $\matL_{\rm pl}$), and consistency regularisation $\matL_{\rm con}$ is imposed in the segmentation task which then generates higher quality pseudo mask $\tilde{y}_i$. The two tasks are modelled using a multi-task network (having two branches  $\hat{g}_{\hat\theta}$ and $f_\theta$ for the two tasks. The weights $\hat\theta$ and $\theta$ are shared between the branches except in the last two blocks). 
The new pseudo masks $\tilde{y}_i$ are used in Stage $3$ (not drawn) which is similar to Stage 2 but $\hat{g}_{\hat\theta}$ is replaced by another network $\tilde{g}_{\tilde\theta}$ (sharing the weights with $f_\theta$ except in the last block).}\label{fig:method}
\end{figure}

With this purpose, we propose a new framework for semantic segmentation in the small data regime.
We approach this problem with a holistic approach, framed as a three-stage self-training technique for semi-supervised learning. Every stage of our framework has its individual purpose: {\textit{Stage 1}} generates initial pseudo-masks (in the sense of pseudo-labelling) by a segmentation network that is trained using the tiny labelled set in a supervised fashion. \textit{Stage 2} is cast as a multi-task model that learns the segmentation (Task 1) and the pseudo-mask statistics (Task 2)  that are used to produce higher quality pseudo-masks. In \textit{Stage 3} the updated pseudo-masks along with the tiny annotation set are used to train the segmentation network, see also Figure \ref{fig:method}. 

Our framework utilises the pseudo-masks and multi-tasking to enforce better predictions on the unlabelled data. It improves the confidence of the pseudo-masks across the stages. This is achieved by introducing a consistency constraint on the unlabelled data and modelling the pseudo masks in a separate task (Task 2). Specifically, our framework does not aim to reproduce individual pseudo-masks in the segmentation task (Task 1) but only taking into account the overall information of pseudo-masks for optimising the segmentation network. On the other hand, the pseudo-masks help to provide additional semantic information apart from the very few ground truth masks in the early learning stages.  We show that the proposed framework achieves the state-of-the-art semi-supervised learning performance using tiny sets of labels. 
In summary, our contributions are as follows. 

\begin{itemize}[noitemsep,topsep=0pt]
\item[$-$] 
We propose a novel self-training framework for learning semantic segmentation at low annotation costs. It generates and improves the pseudo-masks certainty within different stages. 
\item[$-$] 
We introduce a new perspective for semi-supervised semantic segmentation -- a holistic principle. It reduces the uncertainty in the predicted probability for the pseudo-masks using a multi-task model. 
The multi-task model enforces segmentation consistency whilst exploiting the statistical information of the data.
\item[$-$] We carry out extensive experiments to show the performance of the model, and in particular demonstrate that the proposed technique outperforms current semantic segmentation models that rely on very limited ground truth annotations. Ablation studies are conducted to show the importance of the components in the multi-task framework. 
\end{itemize}

\section{Related Work}
Since the seminal work of Long~\cite{long2015fully}, where fully convolutional networks demonstrated potentials to transfer the learnt representation to the task of segmentation, substantial progress has been done using deep supervised techniques for segmentation 
including~\cite{badrinarayanan2017segnet,chen2017deeplab,noh2015learning,romera2017erfnet}.
More recently, sophisticated mechanisms have been combined to overcome the performance limits of existing techniques including  architecture search~\cite{nekrasov2019fast, liu2019auto,zhang2019customizable,lin2020graph}, attention mechanism~\cite{zhao2018psanet,zhang2018context, zhang2019curriculum,huang2019ccnet} and re-designing the principles of several architectures~\cite{yu2018bisenet,li2019dfanet,poudel2019fast, Zhen_2020_CVPR,zhong2020squeeze}. 

Despite the astonishing results reported by these techniques, a major communal drawback is the assumption of a setting under high labelled data regime.  
This has motivated the fast development of techniques relying less on annotations. For example, the family of techniques of weakly-semi supervised segmentation, where weak and strong annotations are combined to drive the performance.  A range of different weak annotations has been used for semantic segmentation. For example, bounding box annotations, scribbles and additional pixel level annotations e.g.~\cite{ahn2018learning,lee2019ficklenet,shimoda2019self, ke2019multi, sun2020mining, ke2020multi,luo2020semi}. We remark that although indeed these techniques seek to ease the labelling process, additional weak labels are used as prior, and therefore,  richer information is used in comparison with solely semi-supervised techniques.
A recent focus of attention is about the developments of deep semi-supervised learning, which is the main topic of this work. In this section, we review the existing techniques in turn. %

\subsection{Deep Semi-Supervised Learning.}
Semi-supervised learning (SSL) has been widely investigated since early developments in the area~\cite{chapelle2009semi}. However, in the last few years there has been a substantial increase of interest on this paradigm, this is particularly because the underpinning theory of SSL has been combined with the power of deep nets reporting impressive results that readily compete with fully supervised techniques. This performance gain has been mainly reported in the context of image classification, where several techniques have been developed e.g.~\cite{laine2016temporal,miyato2018virtual,aviles2019labelled,verma2019interpolation}.
However, \textit{there is a significant difference between image classification and semantic segmentation as the last one involves more dense and complex predictions}.  

Deep SSL for image classification can be achieved using different principles, in which the major success has been achieved through consistency regularisation~\cite{laine2016temporal, tarvainen2017mean, sajjadi2016regularization, miyato2018virtual,berthelot2019mixmatch}. The core idea of this philosophy is that unlabelled samples ($x_u\in \mathcal{D}_u$ denoting the unlabelled set) under induced perturbations, $\delta$, should not change the performance output such that  $f(x_u)= f(x_u+\delta)$. This principle relaxes the clustering assumption of SSL by enforcing an equivalent form of it by pushing the decision boundary to low-density regions. Although the principles applied in image classification can be somehow extrapolated to semantic segmentation, \textit{the inherent gap between tasks prevents semantic segmentation techniques to reach similar high performance than the one reported in classification} (see e.g.~\cite{french2019semi}), and therefore, one needs to rethink the design of deep SSL for semantic segmentation.

\subsection{Deep Semi-Supervised Semantic Segmentation.}
The annotation quality plays a fundamental role in the performance of supervised machine learning techniques. In particular, for the task of semantic segmentation, labelling is overly expensive. For example, a single image from the segmentation benchmark dataset Cityspaces~\cite{cordts2016cityscapes} 
having a resolution of $1024\!\times\!2048$,
involves more than 1M pixel-wise labels. This large amount of pixels makes the annotation task prone to errors, and when designing segmentation methods based on such annotations one needs to account for the  problem of ambiguous pixels. SSL is a way to tackle this problem as the prior relies on a tiny set of labels. Deep SSL for semantic segmentation has only been explored recently in a few works.

Early deep SSL techniques rely on GANs~\cite{goodfellow2014generative}. The authors of~\cite{souly2017semi} propose to enlarge the training set by generating synthetic images to enrich the feature space and strengthen the relationship between unlabelled and labelled samples. Hung et al.~\cite{hung2019adversarial} propose a GAN based technique to differentiate the predicted probability maps from the ground truth segmentations. Similarly, Mittal et al.~\cite{mittal2019semi} propose a two-branch solution composed of: i) a GAN branch that generated per-pixel class labels for the input sample and ii) a multi-label Mean Teacher~\cite{tarvainen2017mean} branch to remove false positive predictions.

More recently, the work \cite{french2019semi} extends CutMix~\cite{yun2019cutmix} to the context of semantic segmentation. 
It applies the principle of strong augmentations, including Cutout, derived from findings in image classification. 
In \cite{ouali2020semi}, consistency is enforced between feature-based, prediction-based and random perturbations. Ke et al.~\cite{ke2020gct} used a flaw probability map along with the extension of the dual student~\cite{ke2019dual} for pixel-wise tasks. An offline self-training scheme based on pseudo-labels with data augmentation has been proposed in~\cite{feng2020semi}  to %
enforce consistency between the labelled and unlabelled sets. The authors of~\cite{mendel2020semi} introduced a two-network approach for semantic segmentation, where the secondary network seeks to further refine the uncertainty predictions by fine-grained error correction.

In the body of literature, a refinement step is commonly used in several deep semi-supervised semantic segmentation techniques.  This is due to the inherent problems of network calibration and the confirmation bias in deep learning e.g.~\cite{guo2017calibration} e.g.~\cite{ke2020gct,mittal2019semi}.
Similar to existing techniques, our approach also deals with the highly uncertain predicted probabilities, through a refinement step, but 
using a carefully designed multi-task framework. More precisely and unlike existing techniques, our holistic approach aims to increase the certainty predictions by enforcing consistency and exploiting the rich statistical information of the data in a multi-task fashion.

\section{Proposed Technique}\label{sec:method}
In this section, we first describe the setting of SSL and the consistency regularisation strategy in our segmentation framework. We then describe the details of the three stages in our training scheme. 

\smallskip
\textbf{Our Deep SSL Setting.} Our holistic technique is framed as a three-stage self-training technique for deep semi-supervised learning. In our setting, we work under the assumption that we have a vast amount of unlabelled data and only a tiny set of annotations. Formally, let the set of labelled samples be denoted by $\qutc{\qut{x_i, y_i} \mid i \in L}$ where $x_i$ is the image and $y_i$ refers to the ground truth segmentation mask corresponding to $x_i$. The set of unlabelled samples are expressed as $\qutc{x_i \mid i \in U}$. We consider the learning problem where only a tiny fraction of the images are labelled, i.e. $|L| \ll |U|$. In what follows, we discuss our approach in detail.

\subsection{Consistency Regularisation with Strong Augmentations}
A key issue in semi-supervised learning is how to compute labels with high certainty for the unlabelled data $x_i \ (i \in U)$. 
In the context of deep learning, prior knowledge has been investigated to better exploit the rich unlabelled data. 

To formulate the consistency regularisation in the setting of SSL segmentation, we define an operator $A: \mathbb{R}^{n\times m\times c} \rightarrow \mathbb{R}^{n\times m\times c}$ that takes the image $x$ to its randomly perturbed versions $Ax$. Common examples of such perturbations include rotation, flipping, translation and their combinations. %
We also define an operator $B: \mathbb{R}^{n\times n \times c'} \rightarrow \mathbb{R}^{n\times n \times c'}$ which maps the pixels of a segmentation mask to a new mask in the same way as $A$ does. The consistency loss is then defined as:
\begin{equation}\label{eq:lcon}
\matL_{\rm con} := \sum_{i \in L \bigcup U} d_c \qutl{ f_\theta\qutl{A x_i}, B f_{\theta'}\qutl{x_i}  },    
\end{equation}
where $f_\theta$ and $f_{\theta'}$ are deep neural network models parametrised by $\theta$ and $\theta'$ respectively, and $d_c$ measures the distance between
$f_\theta\qutl{A x_i}$ and $B f_{\theta'}\qutl{x_i}$. 
In this work, $A$ and $B$ are random operators -- that is, they are different for different samples. 

One key factor of the consistency regularisation \eqref{eq:lcon} is the choice of perturbation operators $A$ and $B$. The minimisation of the cost \eqref{eq:lcon} smooths the network's prediction over a neighbour of the data sample. Therefore, the choice of the operator $A$ reflects the shape and size of the neighbourhood, which affect the segmentation results. In particular, we follow the principle of using strong augmentations as one enforces the perturbations to be more diverse yielding to boost the SSL performance~\cite{xie2019unsupervised}. We then implement the random operators $A$ and $B$ as strong augmenters based on a combination of RandAugment \cite{cubuk2020randaugment} and Cutout \cite{devries2017improved}.

The consistency loss \eqref{eq:lcon} 
is built upon the relationship between two network models. In the literature, the networks $f_\theta$ and $f_{\theta'}$ are often termed as the student model and the teacher model, respectively. The parameter $\theta'$ of the teacher model can be determined in several ways. In the $\Gamma$ model \cite{laine2016temporal}, $\theta'$ is set to $\theta$. In the mean teacher (MT) scheme \cite{tarvainen2017mean},  $\theta'$ is computed as an exponential moving average (EMA) of $\theta$ during the training process. In the MT, as a desired property of the convergence of the student model, $\theta'$ gets closer to the weight $\theta$ of the student as the learning process keeps going \cite{ke2019dual}.
Consequently, the teacher model $f_{\theta'}$ lacks of effective guidance as the number of training steps gets large. %

Motivated by the fact that in consistency regularisation based approaches (e.g. \cite{laine2016temporal, tarvainen2017mean}), the consistency of the models' outputs depends on the coupling of the teacher and the student during the training, we propose a framework that integrates additional guidance via multi-task learning. In the following, we present a three-stage self-training approach that incorporates pseudo-mask statistics to help in learning better, consistent outputs on the unlabelled data. We show that this in return improves the quality of the pseudo-masks within the three stages. 

\begin{table}
\centering
\setlength{\tabcolsep}{3pt}
\caption{Notations used in Stage 2 and Stage 3}\label{tab:notation}
\begin{tabular}{c|c|c|c|c}
\hline
\toprule
& Pseudo-masks & \begin{tabular}{c} Second branch \\ model \end{tabular}  & \begin{tabular}{c} Second branch \\ weight \end{tabular} & Loss function \\
\hline\addlinespace[0.5ex]
Stage 2 & $\hat{y}$ & $\hat{g}$  &  $\hat{\theta}$& $\mathcal{L}^{(2)}$ \\
Stage 3 & $\tilde{y}$ & $\tilde{g}$ &  $\tilde{\theta}$ & $\mathcal{L}^{(3)}$ \\
\hline
\end{tabular}
\end{table}

\subsection{Three-Stage Self-Training Framework}
This subsection contains three core parts of our proposed framework: (i) how we generate the initial pseudo-masks, (ii) a multi-task model to improve the quality of the initial pseudo-masks and (iii) how to propagate the high quality pseudo-masks to the final model. The workflow is displayed in Figure~\ref{fig:method}.

\smallskip
\textbf{Stage 1: Generating Initial Pseudo-Masks.}
In this stage, we use the tiny dataset of labelled images to train a segmentation network $f_\theta$ and generate initial pseudo-masks that have high uncertainty (i.e., low quality pseudo-masks). To do this, we use the following loss function
\begin{equation}\label{eq:lseg}
\matL^{(1)} = \matL_{\rm seg} := \sum_{i \in L} d_s\qutl{f_\theta\qutl{x_i}, y_i}
\end{equation}
where $d_s$ is the cross entropy loss. Once the network is trained, pseudo-masks
are generated from the output of the model $\hat{y}_i := \arg\max f_\theta(x_i)$ where the $\arg\max$ function is performed pixelwise. The main purpose of this stage is to generate 
initial pseudo masks, for which the maximum  predicted  probability is highly uncertain due to the limited amount of ground truth labels.

\smallskip
\textbf{Stage 2: Increasing Certainty for Pseudo-Masks.}
In this stage, the labelled samples, the unlabelled images, and the pseudo masks $\hat{y}_i$ generated from the first stage are used together to produce higher quality pseudo-masks via training a second model.  For ease of presentation, we use the hat symbols (e.g., $\hat{y}_i$) to denote quantities for State 2 (see Table \ref{tab:notation}).
As the masks $\hat{y}_i$ are not accurate, we do not use them to fit the segmentation network, but instead learn to reproduce them in a multi-task model. To do this, we consider two tasks, namely the segmentation task (Task 1) and the auxiliary task (Task 2). Our Task 2 aims to extract the pseudo-mask statistical information from $\hat{y}_i$. 

Our two tasks are framed in a multi-task network  as shown in Figure~\ref{fig:method}.
The first network $f_{\theta}$ is the segmentation network composed of $N$ blocks (i.e., the blue blocks, each block consists of a sequence of layers). Whilst the second network, $\hat{g}_{\hat{\theta}}$ for Task $2$, is composed of $N$ blocks along with the first $N-2$ blocks shared with $f_{\theta}$ (see the green blocks in Figure~\ref{fig:method}), and $\hat \theta$ denotes the shared weights and the own weights of $\hat{g}$. 
For Task 1, the segmentation loss \eqref{eq:lseg} is applied. Additionally, to promote the segmentation accuracy, we include the consistency loss \eqref{eq:lcon} to this task. 
The weights $\theta'$ of the teacher network are updated using the exponential moving average (EMA) of $\theta$ during training \cite{tarvainen2017mean}.
For Task 2 and using pseudo-masks $\hat{y}_i$,  the loss is defined as:
\begin{equation}\label{eq:lpl}
\matL_{\rm pl}  := \sum_{i \in L \bigcup U} d_c \qutl{ \hat{g}_{\hat\theta}\qutl{A x_i}, B \hat{y}_i  }. 
\end{equation}
\textit{We emphasise that the output of $\hat{g}_{\hat\theta}$ is not identical to that of $f_{\theta}$, as the pseudo-masks are not seen by Task $1$. 
This implies that the training process does not enforce the segmentation network $f$ to reproduce individual pseudo-masks which contain errors, but rather the statistics of the pseudo-masks that influence $f$ through back-propagation. By doing so, the pseudo-masks in Task $2$  provide additional guidance to Task $1$ during the optimisation process.} Overall, the loss we use for this stage reads 
\begin{equation}\label{eq:lossstage2}
\matL^{(2)} := \matL_{\rm seg} + \lambda_1 \matL_{\rm con} + \lambda_2 \matL_{\rm pl},
\end{equation}
where $\lambda_1$ and $\lambda_2$ are hyper-parameters balancing the tasks.  
Once the multi-task network is trained, the high quality pseudo-masks are computed by 
$\tilde{y}_i := \arg\max f_\theta(x_i)$. 

\smallskip
\textbf{Stage 3: Propagating Higher Quality Masks Information.}
Finally, in this stage the high quality pseudo-masks $\tilde{y}_i$ from Stage 2 are used to supervise the final models. The high quality pseudo-masks $\tilde{y}_i$ are integrated in a similar way as $\hat{y}_i$ in Stage 2, except that we replace the network $\hat{g}_{\hat{\theta}}$ by $\tilde{g}_{\tilde{\theta}}$ which has $N$ blocks and shares its first $N\!-\!1$ blocks with $f_\theta$ (the network $\hat{g}_{\hat{\theta}}$, however, has only $N\!-\!2$ common blocks with $f_\theta$). This allows the pseudo-mask information to propagate better to $f_\theta$, and hence $\tilde{y}_i$ to be more accurate than $\hat{y}_i$, used in Stage 2.  Here we use the tilde notations (e.g., $\tilde{y}_i$) to denote quantities in Stage 3 (see Table \ref{tab:notation}).
The loss used for this stage is given by
\begin{equation}\label{eq:lossstage3}
\matL^{(3)} := \matL_{\rm seg} + \lambda_1 \matL_{\rm con} + \lambda_2 \matL^{(3)}_{\rm pl},
\end{equation}
where $\matL^{(3)}_{\rm pl}  := \sum_{i \in L \bigcup U} d_c \qutl{ \tilde{g}_{\tilde\theta}\qutl{A x_i}, B \tilde{y}_i  }$.

The overall process of our technique, in which previous individual stages are combined to solve the semantic segmentation problem, is summarised in Algorithm~\ref{alg:threestages}.

\begin{algorithm}[t]
\caption{Our Three-stage Self-training Scheme}\label{alg:threestages}%
\begin{algorithmic}[1]
\footnotesize
\STATE {\coloururl{Input}:}
Labelled samples
$\qutc{\qut{x_i, y_i} \mid i \in L}$  and unlabelled images $\qutc{x_i \mid i \in U}$. The parameters $\lambda_1$ and $\lambda_2$. The random augmentation operators $A$ and $B$. 
\STATE {\textcolor{greenPython}{\# Stage 1:}} 
\STATE \hskip1em Initialise $\theta$ of the segmentation network
\STATE \hskip1em Minimise the loss $\matL^{(1)}$ in \eqref{eq:lseg} for $\theta$
\STATE \hskip1em Compute the initial pseudo masks $\hat{y}_i = \arg\max f_\theta(x_i), \ i \in L\bigcup U$
\STATE {\textcolor{greenPython}{\# Stage 2:}} 
\STATE \hskip1em Initialise $\hat{\theta}$ and reinitialise $\theta$
\STATE \hskip1em With the pseudo-masks $\hat{y}_i$, minimise the loss $\matL^{(2)}$ in \eqref{eq:lossstage2} for parameters $\theta$ and $\hat{\theta}$ of the multi-task network. The weight $\theta'$ of the teacher network is computed as the EMA of $\theta$.
\STATE \hskip1em With the new $\theta$, compute $\tilde{y}_i := \arg\max f_\theta(x_i) \ i \in L\bigcup U$ 
\STATE {\textcolor{greenPython}{\# Stage 3:}}  
\STATE \hskip1em Initialise $\tilde{\theta}$ and reinitialise $\theta$
\STATE \hskip1em Minimise the loss $\matL^{(3)}$ in \eqref{eq:lossstage3} for $\theta$ and $\tilde{\theta}$, where $\tilde{y}_i$ are computed from Stage 2, and $\theta'$ is the EMA of $\theta$
\STATE {\coloururl{Output}:} 
Return the parameter $\theta$ of the segmentation network
\end{algorithmic}
\end{algorithm}

\section{Experimental Results and Discussions}
In this section, we detail the range of experiments that are conducted to evaluate our proposed technique \footnote{The code will be made available at \coloururl{\url{https://github.com/RK621/ThreeStageSelftraining_SemanticSegmentation}}}. 

\subsection{Data Description and Evaluation Protocol}
In the experiments, we use two major benchmark datasets for semantic segmentation. Our first dataset is \textbf{Cityscapes~\cite{cordts2016cityscapes}:} this is an urban scene dataset composed of $2975$, $500$ and $1525$ images for training, validation and test respectively, and it has $19$ classes. The images are of size $1024 \times 2048$. In our experiments, we downsampled all images to the size $512 \times 1024$ following the experimental setting used in \cite{hung2019adversarial, mittal2019semi}. Our second dataset is the augmented \textbf{PASCAL VOC 2012~\cite{everingham2010pascal}:} it is a natural scenes dataset that captures objects from $20$ classes + $1$ background class. The dataset is composed of 10582 and 1449 images for training and validation respectively (parts of the annotations come from \cite{BharathICCV2011}).

To evaluate the performance of the proposed method on the different amounts of labels, we vary the fraction of labelled images for both datasets. Specifically, for the Cityscapes dataset, we carry out experiments on $100$, $372$, $744$ and $1448$ labelled images taken from the whole training set, respectively. We note that all $2975$ training images have ground truth segmentation masks, but some of the segmentation masks are ignored during the training and the corresponding images are treated as unlabelled. Similarly, for the augmented PASCAL dataset, we report results for the models trained on $1/100$, $1/50$, $1/20$ and $1/8$ of the training images (together with the other training images treated as unlabelled), respectively. In both datasets, we follow the same split for the labelled images and unlabelled images as in the work of \cite{mittal2019semi, hung2019adversarial}.

We address our evaluation protocol from both quantitative and qualitative point of views. The former is based on the widely-used metric called the mean Intersection-over-Union
(mIoU) that is used for all our comparisons.
The numerical comparison of our technique is performed against the state-of-the-art methods for deep semantic segmentation: Hung et al.~\cite{hung2019adversarial}, Mittal et al.~\cite{mittal2019semi}, French et al.~\cite{french2019semi},  VAT~\cite{miyato2018virtual}, ICT~\cite{verma2019interpolation}, Feng et al.~\cite{feng2020semi}, Olsson et al.~\cite{olsson2020classmix} and Ke et al.~\cite{ke2020gct}.  The latter   is  based  on  a  visual inspection of the segmentation outputs.

\subsection{Implementation Details}
We provide the training details of our technique for the datasets as well as outline
the reproduced baselines.

\begin{table*}[]
\begin{center}
\caption{Numerical comparison of our technique vs the state-of-the-art models for semantic segmentation on Cityscapes. All results reported are based on the Deeplabv2 network (ResNet-101 backbone). The numerical values are computed as the mIoU metric (in percentage) over different label counts.  '[+MSCOCO]' denotes that the pre-training  considers additionally MSCOCO dataset and only ImageNet otherwise.}
\label{table1:cityscapes}
\resizebox{0.7\textwidth}{!}{
\begin{tabular}{cccccc}
\hline \toprule[1pt]
\cline{2-6}
\multicolumn{1}{c}{\cellcolor[HTML]{EFEFEF}\textsc{Cityscapes}} & \multicolumn{5}{c}{\cellcolor[HTML]{EFEFEF}\# \textsc{Labels}} \\ \hline
\textsc{Technique} & 1/30 (100) & 1/8 (372) & 1/4 (744) & 1/2 (1488) & Full(2975) \\ \hline
DeeplabV2~\cite{chen2017deeplab} & -- & 56.2 & 60.2 & -- & 66.0 \\
Hung et al.~\cite{hung2019adversarial}  & --  & 57.1 & 60.5 & -- & 66.2 \\
Mittal et al.~\cite{mittal2019semi} & -- & 59.3 & 61.9 & -- & 65.8 \\
French (Cutout)~\cite{french2019semi} & 47.21 & 57.72 & 61.96 & -- & 67.47 \\
French  et al. (CutMix)~\cite{french2019semi} & 51.20 & 60.34 & 63.87 & -- & 67.88 \\
\hline
Hung  et al.  [+MSCOCO]~\cite{hung2019adversarial}  & -- & 58.8 & 62.3 & 65.7 & -- \\
Olsson et al. [+MSCOCO]~\cite{olsson2020classmix} & 54.07 & 61.35 & 63.63 & 66.29 & -- \\ \hline  \hline
Ours & \textbf{54.85}  &  \textbf{62.82} &  \textbf{65.80} &  \textbf{67.11} & -- \\ \bottomrule[1pt]
\end{tabular}}
\end{center}
\end{table*}

\begin{table}[]
    \centering
    \setlength{\tabcolsep}{4pt}
    \caption{Results on PASCAL VOC with ImageNet pre-training. Our method is compared to existing techniques for semi-supervised semantic segmentation (all methods use the same Deeplabv2 architectures with ResNet-101 backbone). The numerical values reflect the mIoU metric (in percentage) over different label counts}
    \label{table2:pascalvoc-1}
    \begin{tabular}{cccccc}
    \hline \toprule[1pt]
    \cline{2-6}
    \multicolumn{1}{c}{\cellcolor[HTML]{EFEFEF}\textsc{PASCAL VOC}} & \multicolumn{5}{c}{\cellcolor[HTML]{EFEFEF}\# \textsc{Labels}} \\ \hline
    \textsc{Technique} & 1/100 & 1/50 & 1/20 & 1/8 & Full({\small 10582}) \\ \hline
    DeeplabV2~\cite{chen2017deeplab} & -- & 48.3 & 56.8 & 62.0 & 70.7 \\
    VAT~\cite{miyato2018virtual}   & 38.81 & 48.55 & 58.50 & 62.93 & 72.18 \\
    ICT~\cite{verma2019interpolation}   & 35.82 & 46.28 & 53.17 & 59.63 & 71.50 \\
    Hung et al.~\cite{hung2019adversarial} & -- & 49.2 & 59.1 & 64.3 & 71.4 \\
    Mittal et al. (s4GAN)~\cite{mittal2019semi} & -- & 58.1 & 60.9 & 65.4 &  71.2 \\
    Mittal et al.~\cite{mittal2019semi} & -- & 60.4 & 62.9 & 67.3 &  73.2 \\
    French et al.(Cutout)~\cite{french2019semi} & 48.73 & 58.26 & 64.37 & 66.76 & 72.03 \\
    French et al. (CutMix)~\cite{french2019semi} & 53.79 & \textbf{64.81} & 66.48 & 67.60 & 72.54 \\
    \hline \hline
    Ours & \textbf{55.13} &   62.71 & \textbf{68.23 } &  \textbf{69.36} & -- \\  %
    \cline{1-6}
    \end{tabular}
\end{table}

\begin{table}[]
    \centering
    \caption{Results on PASCAL VOC with MSCOCO Pre-Training. All methods Deeplabv2 architectures with ResNet-101 backbone. The symbol $\dag$ denotes results reproduced by us.}
    \label{table2:pascalvoc-2}
    \setlength{\tabcolsep}{4pt}
    \begin{tabular}{cccccc}
    \hline \toprule[1pt]
    \cline{2-6}
    \multicolumn{1}{c}{\cellcolor[HTML]{EFEFEF}\textsc{PASCAL VOC}} & \multicolumn{5}{c}{\cellcolor[HTML]{EFEFEF}\# \textsc{Labels}} \\ \hline
    \textsc{Technique} & 1/100 & 1/50 & 1/20 & 1/8 & Full({\small 10582}) \\ \hline
    DeeplabV2~\cite{chen2017deeplab} & -- & 53.2 & 58.7 & 65.2 & 73.6 \\
    Hung et al.~\cite{hung2019adversarial} & -- & 57.2 & 64.7 & 69.5 & 74.9 \\
    Zhai et al.~\cite{zhai2019s4l} & -- & -- & -- & 68.65 & 75.38 \\
    Mittal et al. (s4GAN)~\cite{mittal2019semi} & -- & 60.9 & 66.4 & 69.8 &  73.9 \\
    Mittal et al.~\cite{mittal2019semi} & -- & 63.3 & 67.2 & 71.4 &  75.6 \\
    French et al. (CutMix)$\dag$~\cite{french2019semi} & 60.19 & 67.30 & 70.33 & 71.82  & --\\
    Feng et al.~\cite{feng2020semi} & 61.6 & 65.5 & 69.3 & 70.7 & 73.5 \\
    Olsson et al.~\cite{olsson2020classmix} & 54.18 & 66.15 & 67.77 & 71.0 & -- \\
    Ke et al.~\cite{ke2020gct} & -- & -- & -- & 72.14 & 75.73 \\
    \hline \hline
    Ours & \textbf{63.82}  & \textbf{70.77}  &  \textbf{71.90} & \textbf{72.95}  & --  \\ \bottomrule[1pt]
    \end{tabular}
\end{table}

\smallskip
\noindent\textbf{Network architecture and pre-training scheme.} %
In our experiments, we use as our segmentation network,  $f_\theta$, Deeplabv2~\cite{chen2017deeplab} with ResNet-101 backbone~\cite{he2016deep}, which has been also considered in the works of that~\cite{hung2019adversarial, mittal2019semi, french2019semi}. For the experiments on Cityscapes~\cite{cordts2016cityscapes}, the ResNet-101 is pre-trained on ImageNet, whilst for the experiments on PASCAL VOC~\cite{everingham2010pascal}, we use two schemes, the first one of which uses ImageNet pre-training whilst the second one considers additionally MSCOCO~\cite{lin2014microsoft} pre-training for $f_\theta$. 

The networks $\hat{g}_{\hat{\theta}}$ and $\tilde{g}_{\tilde{\theta}}$ are built on the top of $f_{\theta}$. Specifically,  $\hat{g}_{\hat{\theta}}$ (with the same ResNet-101 backbone) has the first $N\!-\!2$ common blocks with $f_\theta$, by sharing its weights of \texttt{conv1}, \texttt{conv2x}, \texttt{conv3x}, \texttt{conv4x} with $f_\theta$, while having its free parameter on \texttt{conv5x} and ASPP layer \cite{chen2017deeplab}. We refer the readers to Table 1 in \cite{he2016deep} for the introduction of \texttt{conv1}, $\cdots$, \texttt{conv5x} blocks of ResNet-101. The network $\tilde{g}_{\tilde \theta}$ shares the blocks \texttt{conv1}, $\cdots$, \texttt{conv5x} with $f_\theta$ while having its own parameter for the ASPP layer. A more detailed description of the structure of $f_\theta$, $\hat{g}_{\hat{\theta}}$ and $\tilde{g}_{\tilde{\theta}}$ is provided in the Appendix. 

We underline that though $\hat{g}_{\hat{\theta}}$ (for Stage 2) and $\tilde{g}_{\tilde{\theta}}$ (for Stage 3) are introduced during the optimisation process, the segmentation network architecture $f_\theta$ remains unchanged and is therefore comparable to the works of \cite{hung2019adversarial, mittal2019semi, french2019semi}. In fact, in the test phase,  $\hat{g}_{\hat{\theta}}$ and $\tilde{g}_{\tilde{\theta}}$ are not included, and therefore the network being evaluated is identical to the DeeplabV2 in e.g., \cite{french2019semi}.

\smallskip
\noindent\textbf{Training Scheme.} 
The training details regarding both datasets are described next. 
For both Cityscape and PASCAL VOC 2012, the overall training  procedure of the proposed method follows Algorithm \ref{alg:threestages}, and the losses for the three stages are listed therein.
In particular, in the consistency loss $\matL_{\rm con}$ \eqref{eq:lcon} and $\matL_{\rm pl}$ \eqref{eq:lpl}, we use the cross entropy for the measure $d_c$. The random operators $A$ and $B$ in the definition of these two losses are implemented as the combination of RandAugment \cite{cubuk2020randaugment} and Cutout \cite{devries2017improved}. At the beginning of each stage, the network parameters, $\theta$, are initialised using the associated pre-trained model (either ImageNet or MSCOCO). In each stage, the network is trained using the Adam optimiser \cite{kingma2014adam} with a learning rate of $3 \! \times 10^{-5}$, and $60,000$ optimisation steps. In each step of the stochastic optimisation, the losses are computed on a minibatch. %
More specifically, each minibatch consists of Sub-batch 1 whose images are randomly selected from $L \bigcup U$ and Sub-batch 2 whose images are from $L$. The loss $\matL_{\rm con}$ and $\matL_{\rm pl}$ are averaged over both sub-batches, while the loss $\matL_{\rm seg}$ is averaged over Sub-batch 2 only. In our implementation, we do not perform strong augmentation on Sub-batch 2, i.e., $A$ and $B$ are identity operators (see the loss $\matL_{\rm seg}$  defined in \eqref{eq:lseg}). 

For Cityscapes, all images are downsampled into $512 \! \times \! 1024$ throughout the experiments. During the training phase, all images are cropped into $256\!\times \! 512$ before feeding to the network. Random horizontal flipping is applied to augment the dataset. The same setting was used in the works of \cite{hung2019adversarial, mittal2019semi}. To apply our three-stage self-training method, we use a minibatch size of $5$. We consider different labelled images counts using  $|L|=100$ ($1/30$ of the whole set of images), $372$, $744$ and $1448$ respectively.
The size of Sub-batch $1$ is $1$ for the cases $|L|=100$, $372$ and $744$, while this number is increased to $3$ when $|L|$ increases to $1448$. 
For all cases, unless specified otherwise, the parameters $\lambda_1$ and $\lambda_2$ are set to $1/2$. 

For PASCAL VOCC 2012, the images are cropped into sizes of $321\! \times \! 321$ during the training. Augmentations, including random horizontal flipping and random scaling, are applied, which follows the setting of semi-supervised segmentation benchmarks \cite{hung2019adversarial, mittal2019semi}. We trained the network using $1/100$, $1/50$, $1/20$ and $1/8$ labelled images respectively. The size of the minibatch is set to $10$. 
In each minibatch, the size of Sub-batch $1$ is $1$ for the setting with $1/100$, $1/50$, $1/20$ labels, and this number is increased to $2$ when the fraction of labelled images increases to $1/8$.
For this experiment, the parameter $\lambda_1$ is set to $1/4$, $1/4$, $5/4$, $5/8$ for the $4$ different amounts of labels respectively, and $\lambda_2 = \lambda_1$.

\subsection{Results and Discussions}
In this section, we discuss the findings drawn from our numerical and visual results.

\begin{figure}[t!]
    \centering
    \begin{tabular}{c|c}
        \includegraphics[width=0.48\linewidth, trim=0 12 0 0, clip]{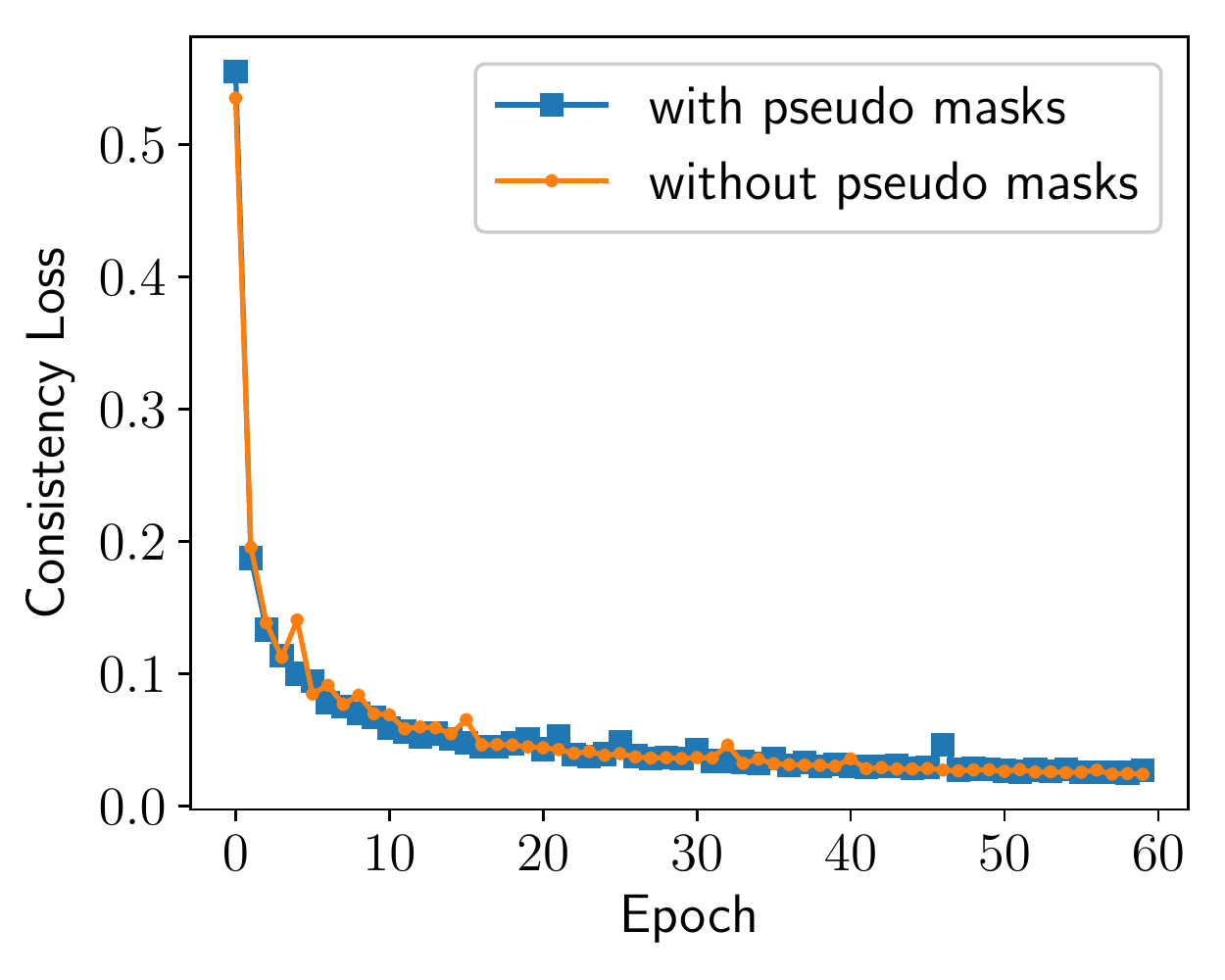} 
         &  
         \includegraphics[width=0.48\linewidth, trim=0 12 0 0, clip]{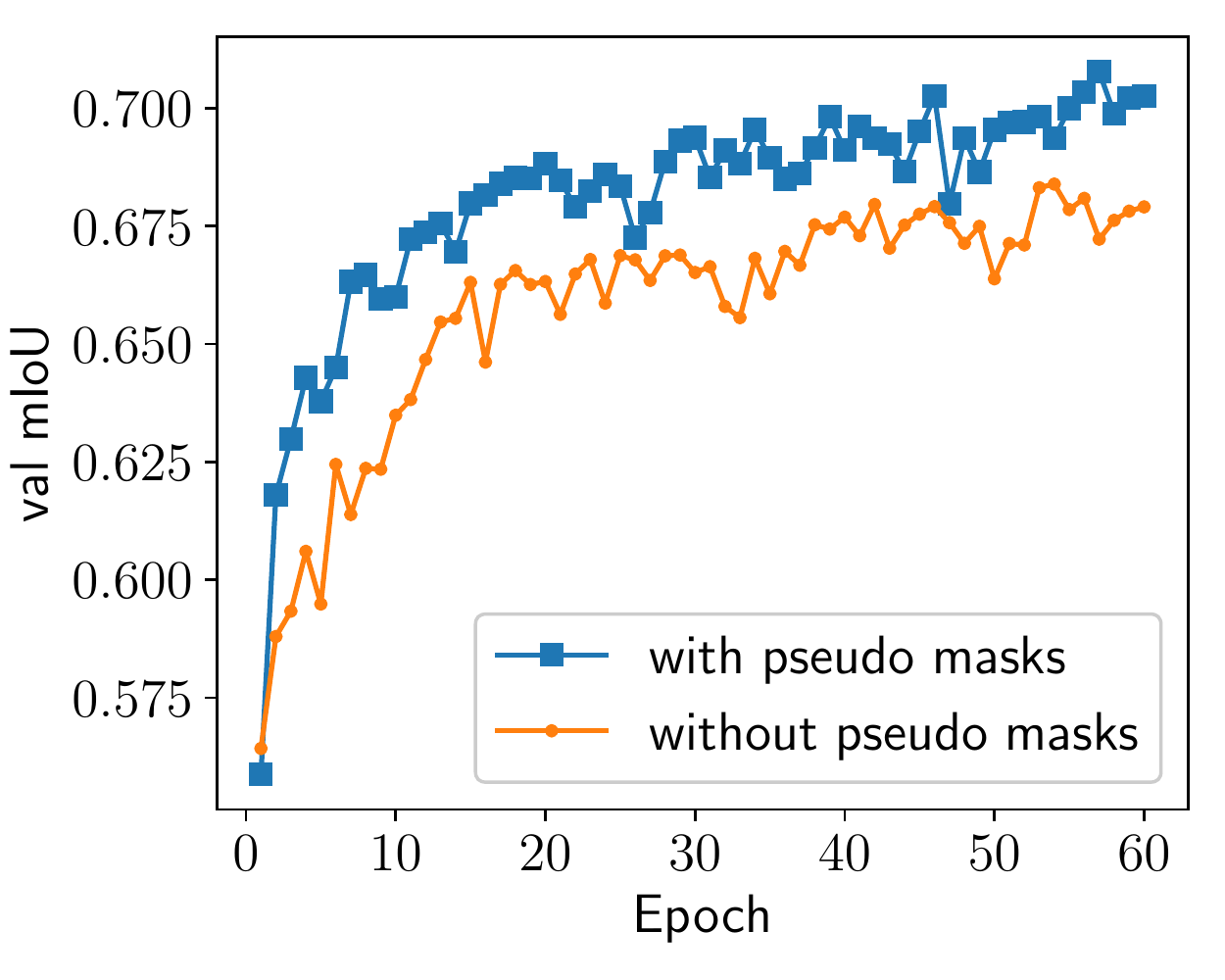} 
    \end{tabular}
    \caption{\small Optimisation with pseudo-masks (PASCAL VOC (212))}
    \label{fig:ImproveOpt}
\end{figure}

\smallskip
\noindent\textbf{Influence of the pseudo-masks}
We start by looking at how the pseudo-masks improve SSL. To this end, we compare the consistency of the segmentation as well as the validation accuracy between the case with pseudo masks and the case without pseudo-masks. For the latter, the loss term $\matL_{\rm pl}^{(3)}$ is removed from $\matL^{(3)}$ (see \eqref{eq:lossstage3}) in Stage $3$. We consider the PASCAL VOC (212 labels) and the results are plotted in Figure \ref{fig:ImproveOpt}. The left part of Figure \ref{fig:ImproveOpt} shows the value of the consistency loss $\matL_{\rm con}$ over the training epochs, and the value decreases in a similar way for both cases (with or without the pseudo masks). This indicates that the segmentation outputs have increasing consistently during the training regardless the use of the pseudo-masks. However, on the right part of Figure \ref{fig:ImproveOpt}, we observe a boost in the validation mIoU of the segmentation outputs especially at the first few epochs if the pseudo-masks are used. This means that the semantic information in the pseudo-masks help to find a better segmentation, while the consistency regularisation is enforced.

\smallskip
\noindent\textbf{Comparison of our Technique vs SOTA Models.} We start by comparing our technique against existing semi-supervised semantic segmentation models for both datasets.  

\textit{Performance Comparison using Cityscapes.} 
The results are reported in Table~\ref{table1:cityscapes}. We report two case studies in this table. The first case uses
ImageNet as pre-training (see the upper part of Table~\ref{table1:cityscapes}), which is used by our approach. From this comparison, we can observe that our approach significantly improves over all compared techniques and for all labels counts. Interestingly, we  observe that only with $1/4$ and $1/2$ of labels, we can reach the performance that only is reachable for the compared techniques using the full labelled set. The second one is against techniques that additionally use MSCOCO (see techniques with the tag '[+MSCOCO]') in the pre-training stage, 
we observe that our technique without such additional data is able to outperform these methods.

\begin{figure*}[!t]
\centering
\includegraphics[width=1\textwidth]{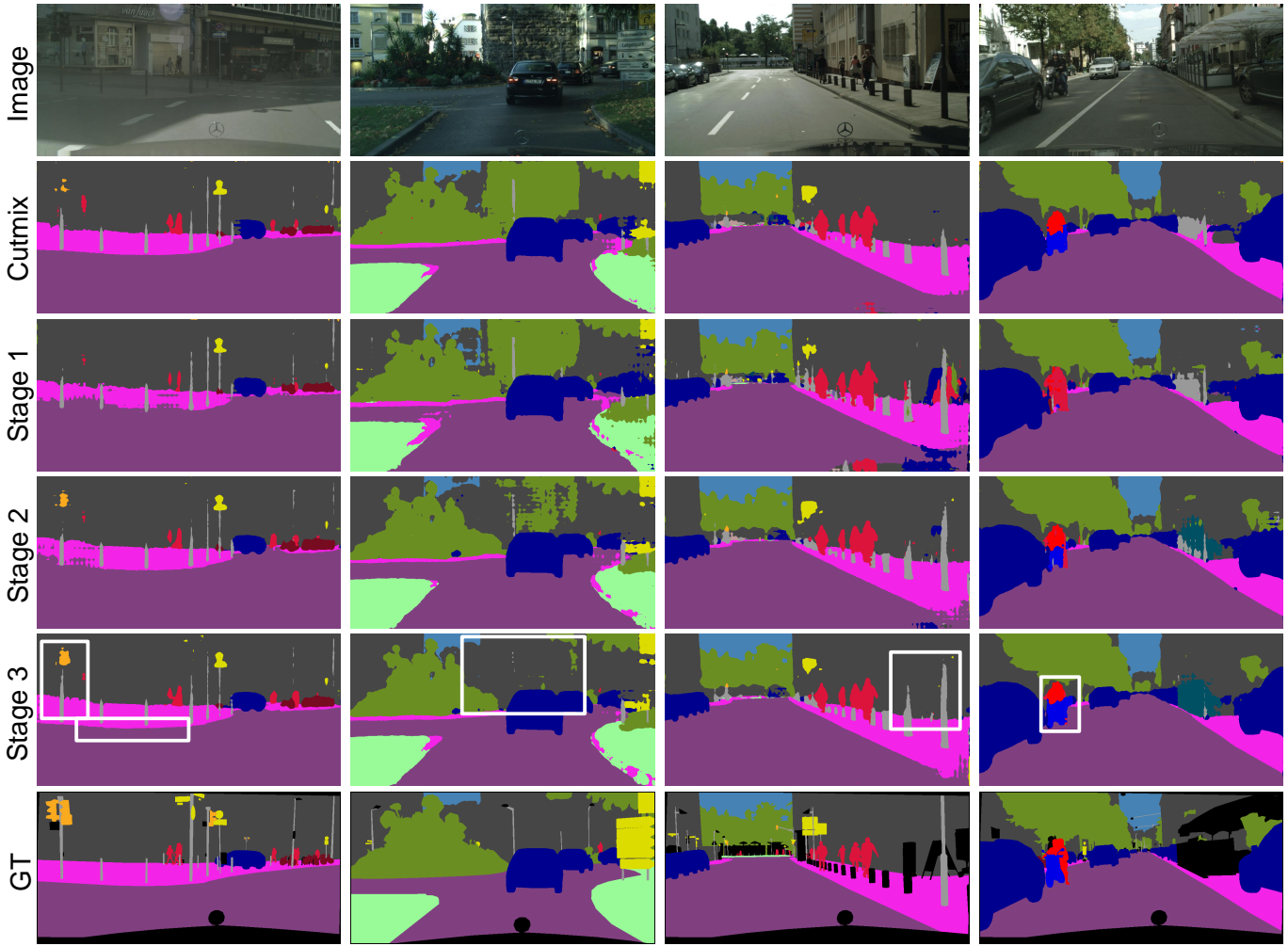}
\caption{Visual comparison of our technique vs CutMix~\cite{french2019semi}. The first row displays the sample images. The remaining rows represent the segmentation outputs for:  CutMix, Stage 1 (Ours), Stage 2 (Ours), Stage 3 (Ours), and the ground truth (GT) respectively. 
The white rectangles highlight the fine details that are captured by the last stage of our method. 
}
\label{fig::suppFig1}
\end{figure*}

A visual comparison of the segmentation outputs of each stage of the proposed method, as well as that of CutMix \cite{french2019semi}, are presented in Figure~\ref{fig::suppFig1}. The segmentation errors %
are significantly reduced from Stage 1 to Stage 3, and our method shows better results than CutMix. In a closer inspection at these segmentation outputs, one can see that our technique was able to segment fine structures -- for example the poles, traffic lights and trees which are highlighted using white rectangles. By contrast, the results of that~\cite{french2019semi} failed to retrieve many of such relevant regions.

\begin{table}[t!]
\begin{center}
\caption{Ablation study on Cityscapes (CS) and PASCAL VOC (P VOC) datasets. We refer to S1, S2 and S3 as each corresponding stage of our technique. Moreover, we indicate consistency regularisation as CR, Task 2 as T2 and strong augmentation as SA. Finally, the option N2 means the use of $\tilde{g}_{\tilde\theta}$ in Stage 3 (otherwise $\hat{g}_{\hat{\theta}}$ is used in Stage 3 instead).
}\label{tab:ablation}
\resizebox{0.49\textwidth}{!}{
\setlength{\tabcolsep}{5pt}
\begin{tabular}{ccccccc|cc}
\hline \toprule[1pt]
S1 & S2 & S3 & CR & T2 & SA & N2 &  \cellcolor[HTML]{EFEFEF}\textsc{CS} (100) & \cellcolor[HTML]{EFEFEF}\textsc{P VOC} (212)  \\ \hline
\checkmark & & & NA & NA & NA & NA & 46.91 & 55.20 \\
\checkmark & \checkmark &  & \checkmark & \checkmark & \checkmark & {NA} & 53.25 &  68.50 \\
\checkmark & \checkmark &  \checkmark &    & \checkmark & \checkmark & \checkmark  & 50.2 & 59.82 \\
\checkmark & \checkmark &  \checkmark & \checkmark  &  & \checkmark & {NA} & 52.63 & 68.21 \\
\checkmark & \checkmark &  \checkmark & \checkmark & \checkmark &  & \checkmark & 47.96 & 64.37 \\
\checkmark & \checkmark &  \checkmark & \checkmark & \checkmark & \checkmark &  & 53.80 & 69.52 \\
\hline
\checkmark & \checkmark & \checkmark  & \checkmark & \checkmark & \checkmark & \checkmark  & 54.85 & 70.77 \\
\hline

\cline{1-6}
 \bottomrule[1pt]
\end{tabular}
}
\end{center}
\end{table}

\begin{table}[t!]
    \centering
    \caption{The mIoU (in percentage): with/without EMA during training}\label{tab:EMA}
    \begin{tabular}{c|cc|cc}
    \hline
    \hline
        \multirow{2}{*}{EMA} &  \multicolumn{2}{c|}{\cellcolor[HTML]{EFEFEF}\textsc{Cityscapes} (100)}  & \multicolumn{2}{c|}{\cellcolor[HTML]{EFEFEF}PASCAL VOC (212)}  \\
        \cline{2-5} 
        & S2 & S3 & S2 & S3 \\
        \hline 
         yes & 53.25 & 54.85 & 68.50 & 70.77 \\
         \hline 
          no & 51.68 & 53.37 & 64.88 & 66.83  \\
         \hline
         \hline 
    \end{tabular} 
  \end{table}

\textit{Performance Comparison using PASCAL VOC.}
We further investigate the performance of our technique by a set of comparisons using the PASCAL VOC dataset. 
Similarly than in the previous set of comparisons, we have two cases with and without (i.e. only ImageNet) MSCOCO pre-training. We begin the comparison using the case without MSCOCO and the results, measured in mIoU, are reported at Table~\ref{table2:pascalvoc-1}. In a closer look at the results, we observe that our technique is consistent 
with its performance behaviour on Cityscapes as our technique substantially improves over the compared algorithmic techniques for all label counts. The only exception is for \cite{french2019semi} in the case of  $1/50$ labelled data. However, one can see that our technique substantially improves (from 62.71\% to 70.77\% in mIoU) when using MSCOCO for this particular case. An interesting case is when using VAT and ICT techniques, which drawn directly from image classification. From the numerical results, we observe that these two techniques report low performance which highlights the substantial gap in modelling hypothesis between classification and semantic segmentation, where  the second one requires more dense and complex annotations.

The results on PASCAL for the case with MSCOCO pretraining are displayed at Table~\ref{table2:pascalvoc-2}. By inspection, we observe that the performance of our technique is prevalently outperforming all compared techniques for all labels counts. Particularly, our method reports large improvement for this case up to $\sim 33\%$, most notably, when a lower label regime is used i.e. $1/100, 1/50, 1/20$.
Unlike Cityscapes where the performance gain using MSCOCO pretraining is small, one can observe that for the case of PASCAL VOC the improvement is significant. This is mainly because the samples content  from MSCOCO is similar than PASCAL VOC. This allows to further enrich the data generalising better the performance for this case. 

\begin{figure*}
    \centering
    \setlength{\tabcolsep}{4pt}
    \begin{tabular}{cccccc}
    \includegraphics[width=0.15\linewidth]{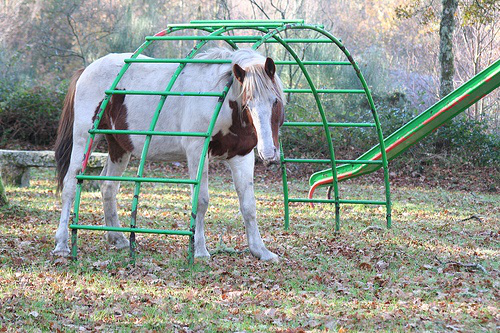} &
    \includegraphics[width=0.15\linewidth]{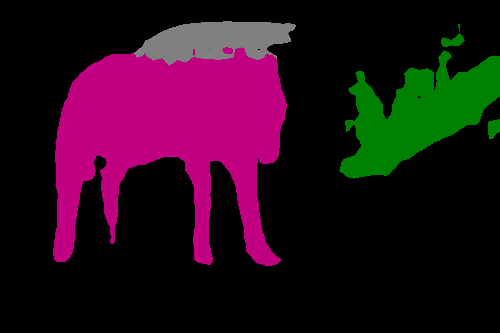} & 
    \includegraphics[width=0.15\linewidth]{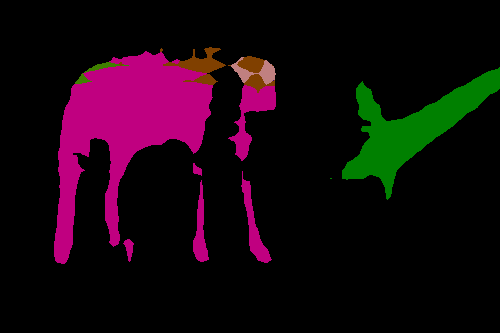} &
    \includegraphics[width=0.15\linewidth]{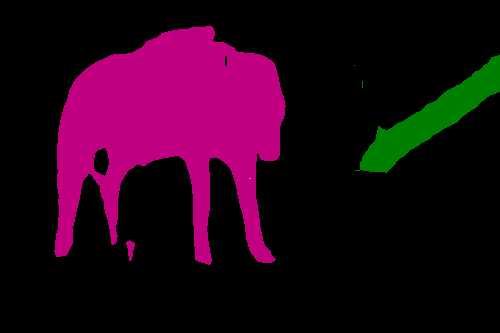} &
    \includegraphics[width=0.15\linewidth]{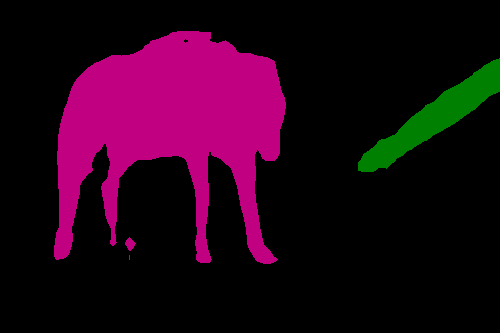} &
    \includegraphics[width=0.15\linewidth]{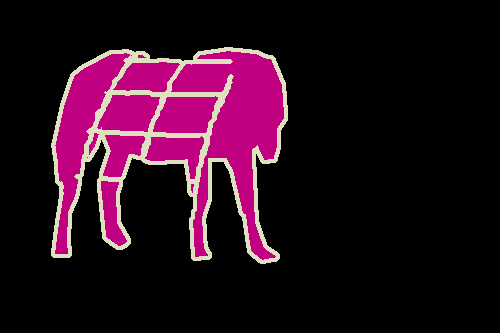} 
    \\
    \includegraphics[width=0.15\linewidth]{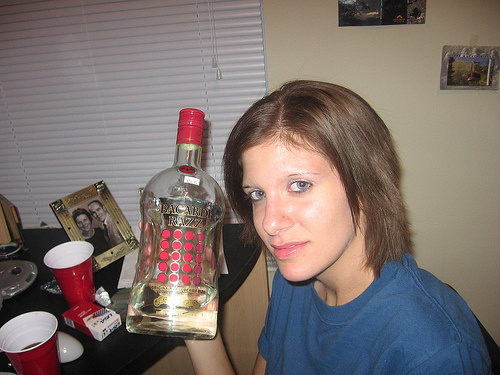} &
    \includegraphics[width=0.15\linewidth]{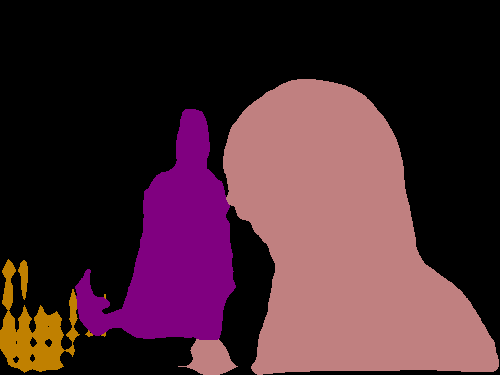} & 
    \includegraphics[width=0.15\linewidth]{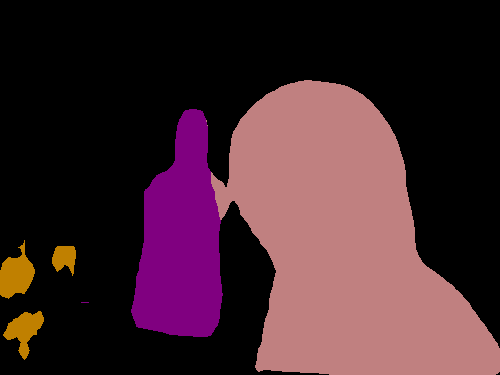} &
    \includegraphics[width=0.15\linewidth]{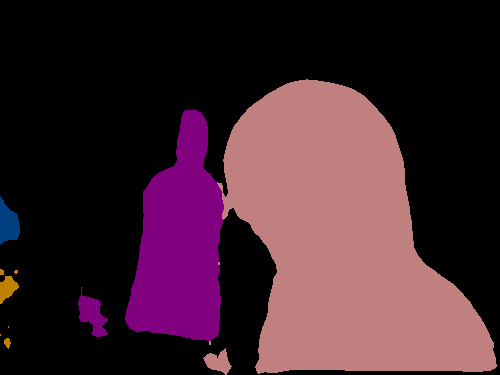} &
    \includegraphics[width=0.15\linewidth]{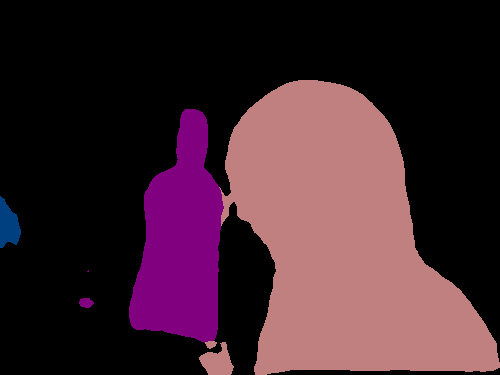} &
    \includegraphics[width=0.15\linewidth]{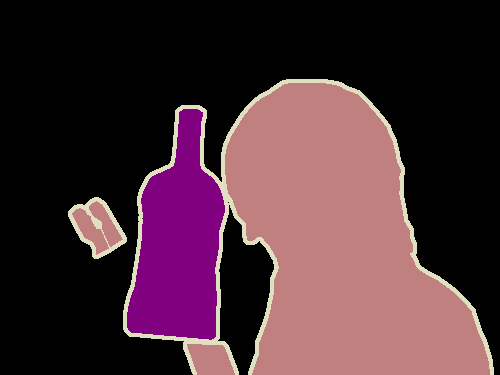} 
    \\
    \includegraphics[width=0.15\linewidth]{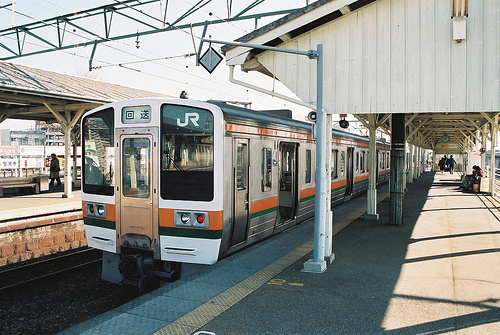} &
    \includegraphics[width=0.15\linewidth]{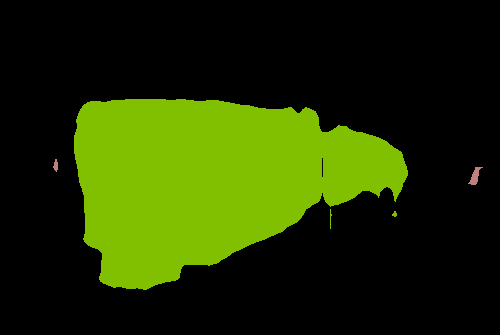} & 
    \includegraphics[width=0.15\linewidth]{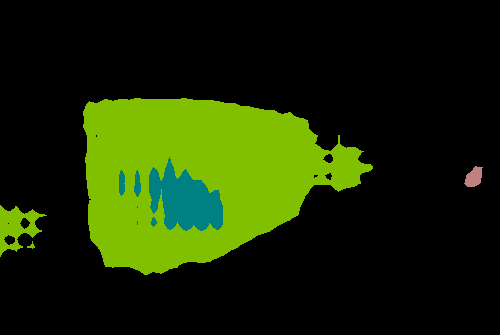} &
    \includegraphics[width=0.15\linewidth]{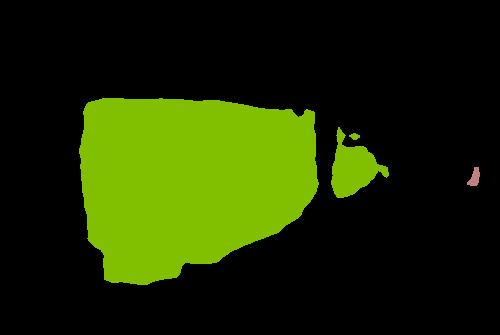} &
    \includegraphics[width=0.15\linewidth]{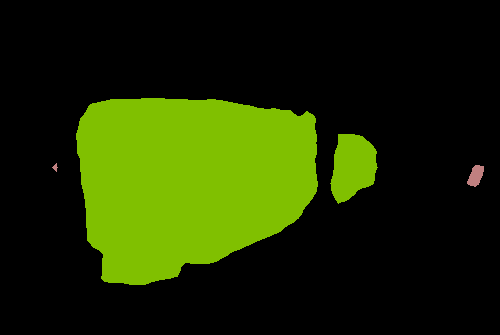}  &
    \includegraphics[width=0.15\linewidth]{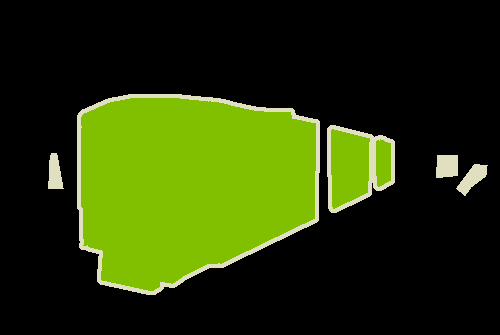}  
    \\
    \includegraphics[width=0.15\linewidth]{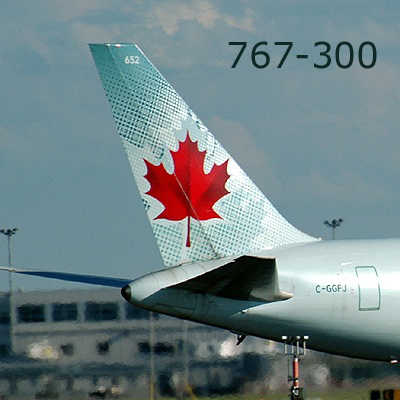} &
    \includegraphics[width=0.15\linewidth]{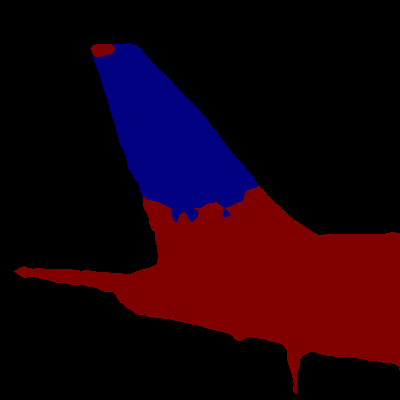} & 
    \includegraphics[width=0.15\linewidth]{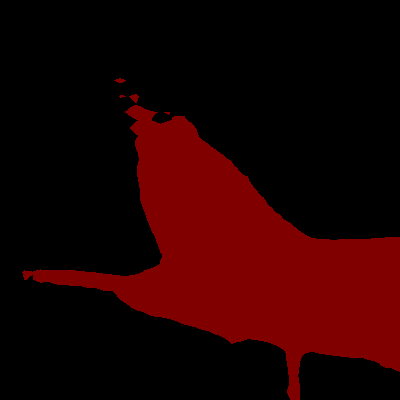} &
    \includegraphics[width=0.15\linewidth]{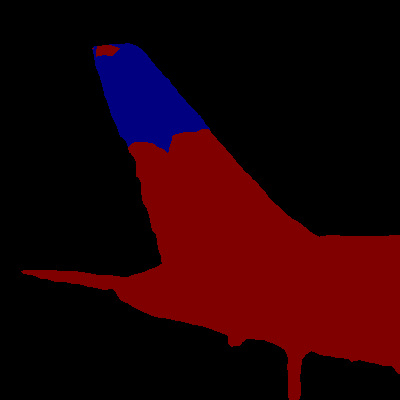} &
    \includegraphics[width=0.15\linewidth]{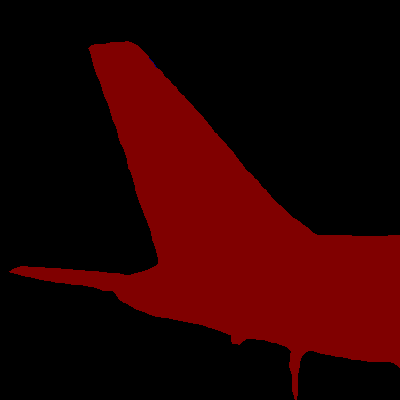}  &
    \includegraphics[width=0.15\linewidth]{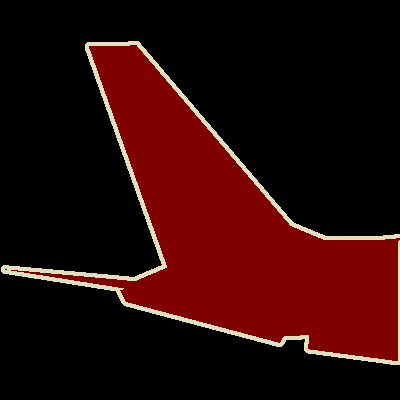}  
    \\
    \includegraphics[width=0.15\linewidth]{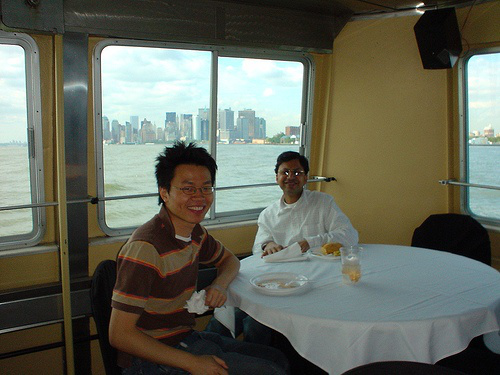} &
    \includegraphics[width=0.15\linewidth]{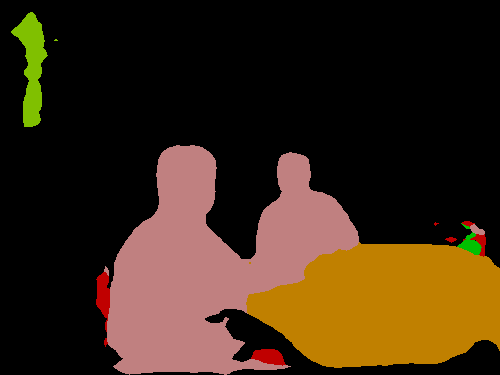} & 
    \includegraphics[width=0.15\linewidth]{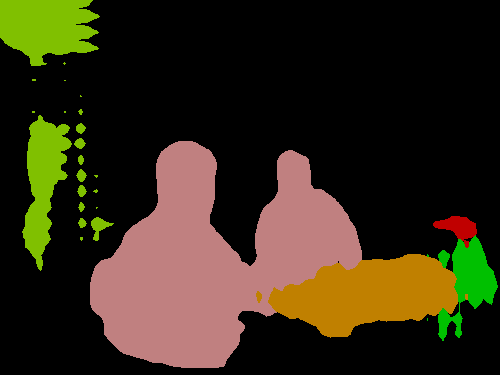} &
    \includegraphics[width=0.15\linewidth]{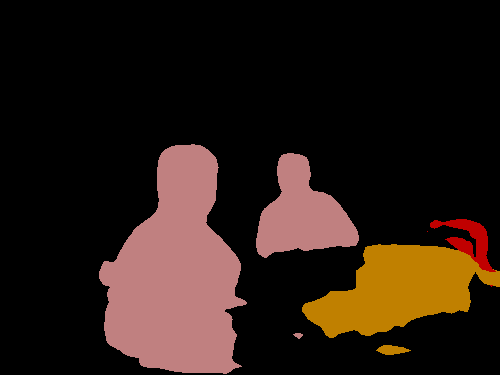} &
    \includegraphics[width=0.15\linewidth]{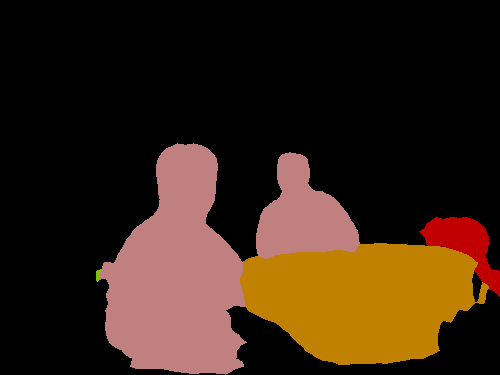}  &
    \includegraphics[width=0.15\linewidth]{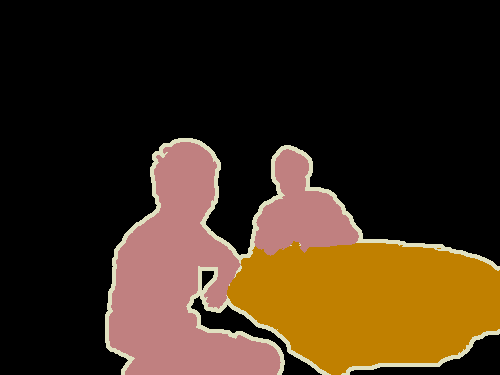}  
    \\
    \includegraphics[width=0.15\linewidth]{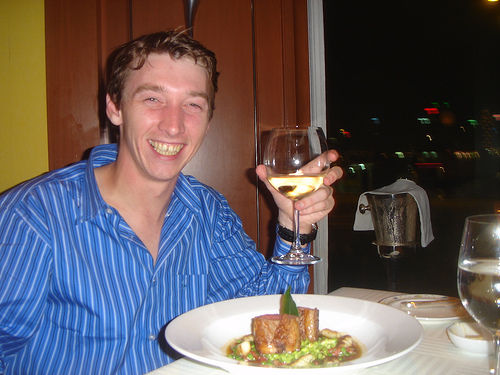} &
    \includegraphics[width=0.15\linewidth]{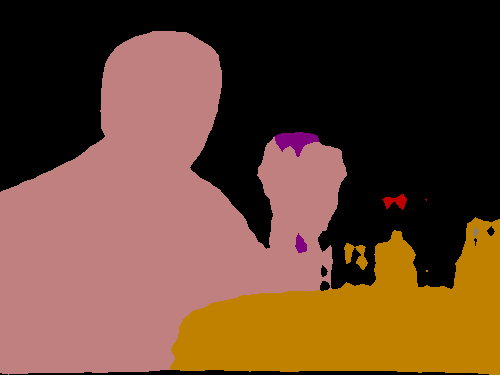} & 
    \includegraphics[width=0.15\linewidth]{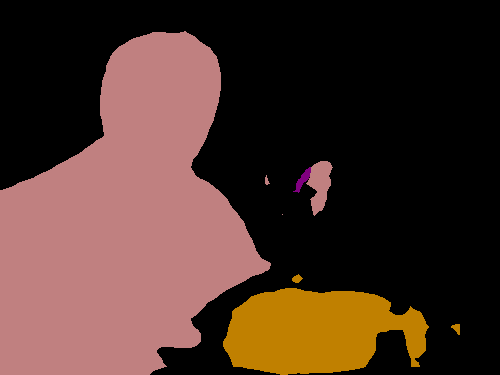} &
    \includegraphics[width=0.15\linewidth]{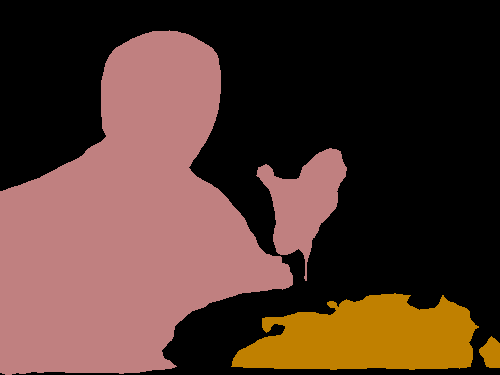} &
    \includegraphics[width=0.15\linewidth]{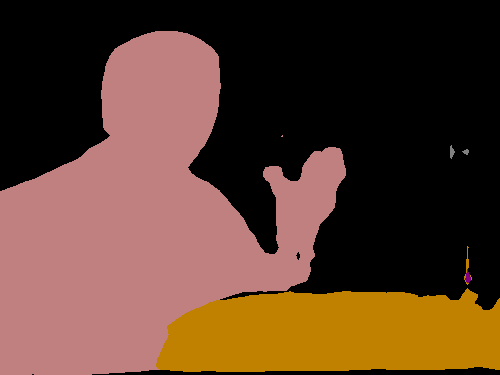}  &
    \includegraphics[width=0.15\linewidth]{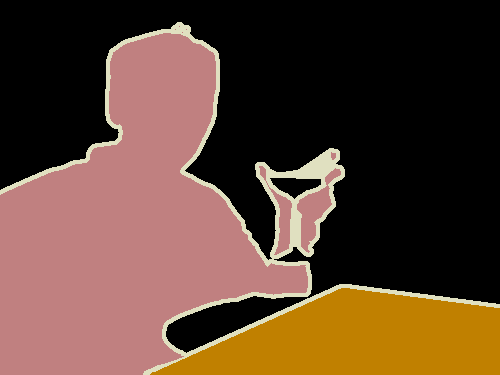}  
    \\
    Image & CutMix \cite{french2019semi} & Stage 1 & Stage 2 & Stage 3 & Ground truth \\
    \addlinespace[1.0ex]
    \end{tabular}
    \caption{Visual comparison of the different stages of our technique vs a SOTA technique on a selection of images from PASCAL VOC validation sets. The models are trained using $212$ ground truth segmentation masks. The first column displays the sample input whilst the 2-5 columns are the segmentation outputs of CutMix and Ours (Stage 1-3), respectively. 
    }
    \label{fig:pascal212}
\end{figure*}

Figure \ref{fig:pascal212} compares the segmentation outputs of the three stages, along with that of CutMix \cite{french2019semi} and the ground truth annotations, for images taken from the PASCAL validation set. As shown in the figure, more details of the foreground objects are captured from Stage 1 to Stage 3, and the final segmentation results are more accurate at the object boundaries than the segmentation by CutMix.

\begin{figure}[t!]
    \centering
    \begin{tabular}{cc}
    \includegraphics[width=0.42\linewidth, trim=10 10 10 10, clip]{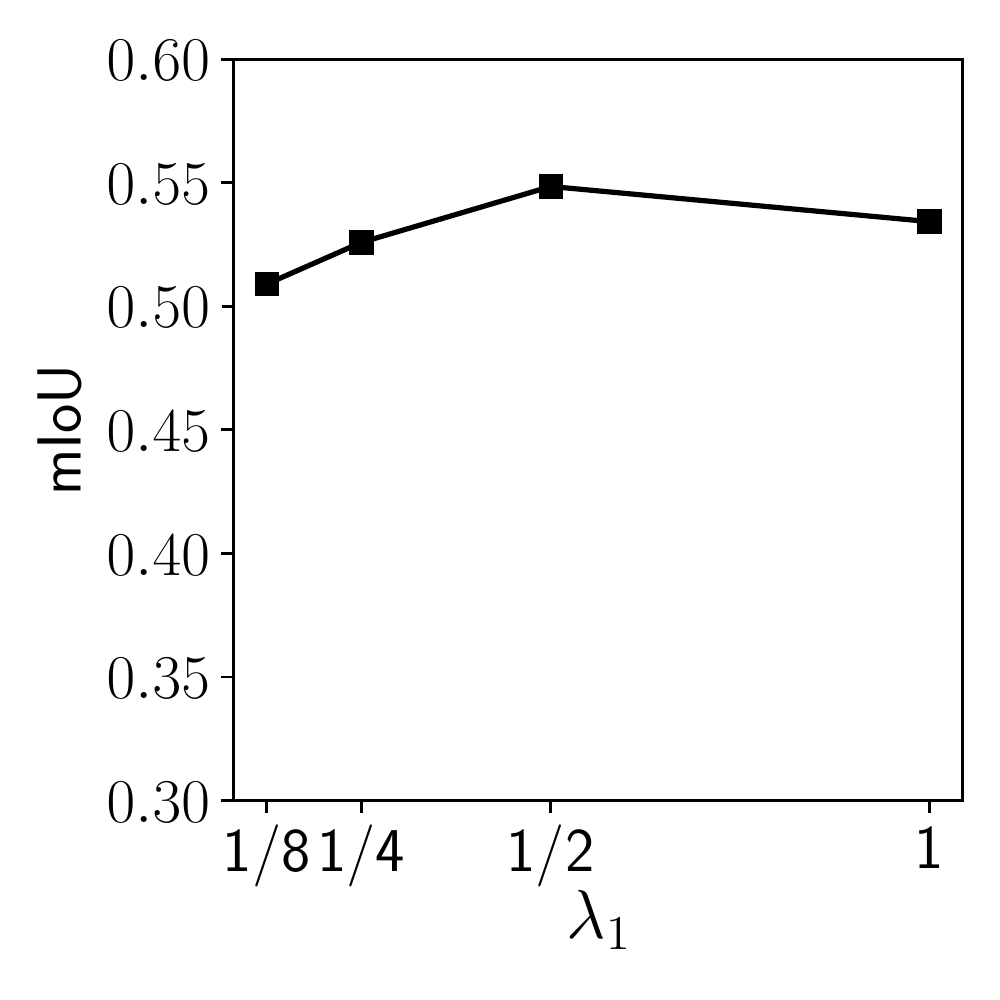} 
    & 
    \includegraphics[width=0.42\linewidth, trim=10 10 10 10, clip]{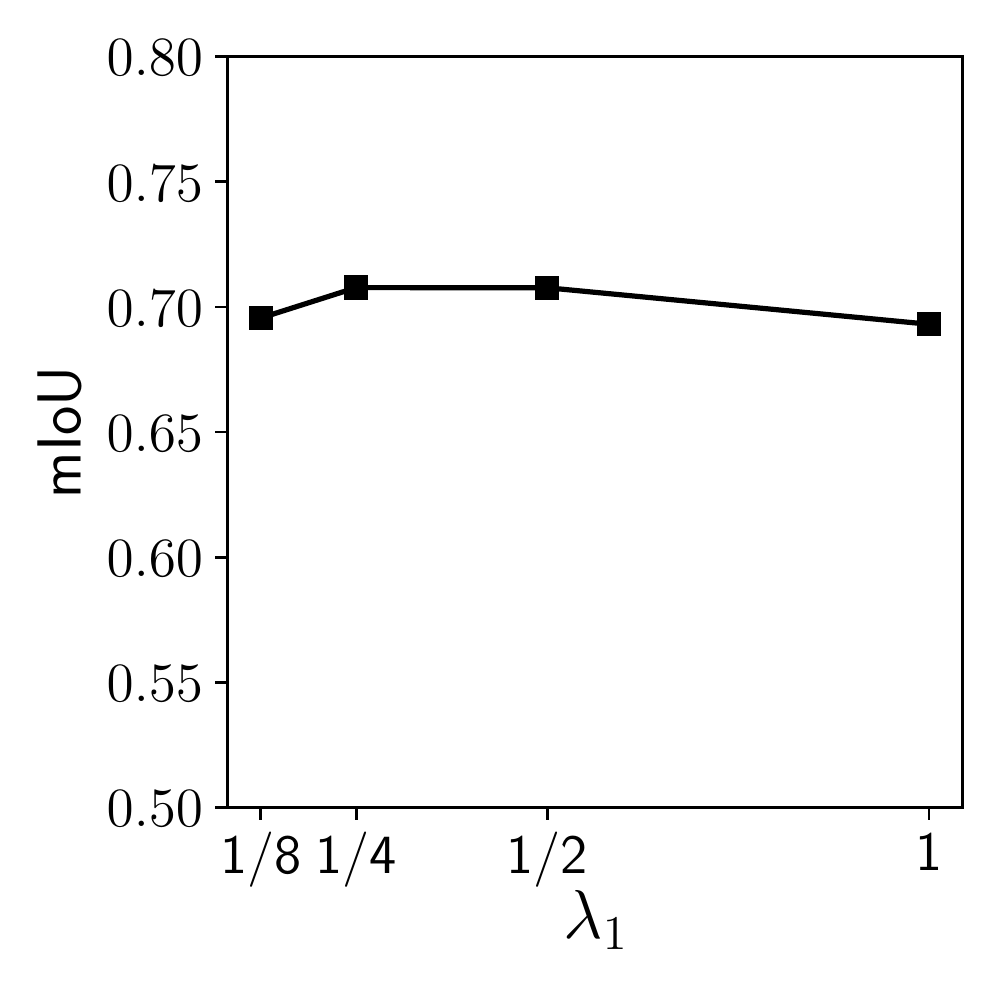}\\
    (a). Cityscapes & 
    (b). PASCAL 
    \end{tabular}
    \caption{{Behaviour curves of } the mIoU vs the values of $\lambda_1$.}
    \label{fig:lambda1}
\end{figure}

\smallskip
\noindent\textbf{Ablation study on the tasks and the augmentations.}
Results of ablation study for Cityscapes (with 100 labelled images and ImageNet pretraining) and PASCAL VOC (with 212 labelled images and MSCOCO pretraining) are reported in Table \ref{tab:ablation}. Starting from Stage 1 (S1), the performance is substantially improved by including the second stage (S2). Stage 3 (S3) further improves the mIoU by $1.6\%$ for Cityscapes and $2.2\%$ for PASCAL. A significant drop of the score ($\geq\!4$\%) appears when the consistency regularisation (CR) $\matL_{\rm con}$ is not included, which suggests that CR helps to improve the segmentation accuracy in addition to the pseudo-masks. A loss of around $2$\% in the mIoU is observed when Task 2 (T2) is dropped, indicating the importance of Task 2 (using the pseudo-masks). Strong augmentation (SA) is an important component of the framework, as excluding SA decreases the score by around $6.5\%$ for both datasets. Finally, changing the network $\tilde{g}_{\tilde \theta}$ (N2) for Stage 3 into $\hat{g}_{\hat \theta}$ (used by Stage 2) results in about $1\%$'s performance loss, which indicates that $\tilde{g}_{\tilde \theta}$ (sharing more blocks with $f_\theta$) allows better propagation of the pseudo-mask information to $f_\theta$. 

We also performed an ablation study on the influence of the exponential moving average (EMA), and the results are reported in Table \ref{tab:EMA}. Without the EMA, the segmentation quality degrades resulting in a decrease in performance in terms of mIou of
$1.5\%$ for Cityscapes (100 labels), 
while $4\%$ for the Pascal VOC dataset (212 labels).

\smallskip
\noindent\textbf{The weights $\lambda_1$ and $\lambda_2$ of the losses.} 
We study how the choice of the weights $\lambda_1$ and $\lambda_2$ affects the performance of the proposed method. To do this, we report the results for both datasets over different values of the weights in Figure \ref{fig:lambda1}. In this test, for Cityscapes, $100$ ground truth segmentation masks are used for training, and the segmentation network DeeplabV2 is pre-trained using ImageNet. For the augmented PASCAL dataset, we use $2\%$ labelled images and MSCOCO pre-training. We set equal weights for the consistency loss and the pseudo-masks loss term, i.e., $\lambda_2 = \lambda_1$. 

From the results in Figure~\ref{fig:lambda1} (a), the best performance for the Cityscapes dataset is reached at $\lambda_1 = 1/2$, and the mIoU decreases by around $2$\% and $1.5\%$ when $\lambda_1$ is changed to $1/4$ and $1$, respectively. 
For the augmented PASCAL dataset, the highest mIoU ($0.708$) is archived at $\lambda_1 = 1/4$ and $\lambda_1=1/2$ (See Figure \ref{fig:lambda1} (b)). We observe a slightly drop in the mIoU (around $1\%$) when $\lambda_1$ is set to $1/8$~or~$1$.

\smallskip
\noindent\textbf{Comparing to representation learning.}
Our framework enforces better predictions on unlabelled data through a multi-task and multi-stage process. Alternatively, one could integrate representation learning tasks into multi-task frameworks, such as contrastive learning (e.g., MoCo \cite{he2020momentum}) or image classification tasks (given additional image-level labels).
The learned representation is shared with the segmentation task and helps to improve the segmentation predictions on the (unlabelled) data where ground truth segmentation masks are unavailable. 
These representation learning tasks, however, do not necessarily enforce consistent predictions in pixel-level (needed for good segmentation) because the contrastive loss or classification loss does not have an implication of how the position of a pixel in one image is correlated to the same pixel in another image.
In contrast, our framework fully leverages the pixel-wise correspondence of an image and its augmented versions (see the losses \eqref{eq:lcon} and \eqref{eq:lpl}) and explicitly constrains the model to predict segmentations that are consistent over the different augmented images.

\smallskip
\noindent\textbf{The multi-task network and extensions.}
The multi-task networks for Stage $2$ and Stage $3$ play an important role in balancing the uncertainty in the pseudo-masks and the consistency of the prediction. This is crucial especially when the labelled data is small and the initial pseudo-masks  reflect low accuracy. %
As described in Section \ref{sec:method}, the network $\hat{g}_{\hat{\theta}}$ is constructed by reusing the first $N\!-\!2$ blocks of the segmentation network $f_\theta$, while having its independent parameters for the last $2$ blocks. The results displayed in this work are based on Deeplabv2, but we remark
that the multi-task network can be constructed in a similar way if other segmentation networks are adopted. %

The proposed method consists of three stages, and the quality of the pseudo masks is improved between the stages. In our experiments, we observe that the three stages reflect a good trade-off between performance and computation. For example, when a fourth stage is included, the mIoU is improved by 0.12\% (at 70.89\%) for PASCAL VOC (212 labels) and 0.36\% (at 55.21\%) for Cityscapes (100 labels), respectively. Also, as shown in Table \ref{table1:cityscapes}, Table \ref{table2:pascalvoc-1} and Table \ref{table2:pascalvoc-2}, the three stage scheme archives SOTA performance for semi-supervised semantic segmentation. However, this does not restrict an extension of the main idea to further stages, because the improvements from one stage to the next one are encoded in the pseudo-masks, which are readily usable if more stages are included.

\section{Conclusion}
In this work, we propose a three-stage self-training method for semi-supervised semantic segmentation. %
Our framework enforces consistency of labels via a moving teacher network and introduces an auxiliary task to propagate the pseudo-mask information (which are stationary within each stage of the training) to the student network during the optimisation process. 
We demonstrate that our framework effectively improves the quality of pseudo-masks, during the learning process using a multi-task model, given only a very small amount of ground truth segmentation masks. The proposed method reaches state-of-the-art semi-supervised segmentation results on two major benchmark datasets.  
Interesting extensions of the proposed method can be derived by combining our approach with additional pseudo-mask enhancement techniques, including uncertainty weighting and warm-up. Another research line is to investigate how multi-task can intertwine with representation learning techniques such as \cite{he2020momentum}, and additionally, how to integrate our auxiliary tasks and classification tasks.

\appendix[Further Details of the Network Architecture]
In this section, we provide the specification of the multi-task network used in Section 4 of the paper. 

The multi-task network is adapted from a segmentation network. In particular, each of the branches $f_\theta$, $\hat{g}_{\hat\theta}$, and $\tilde{g}_{\tilde{\theta}}$ has $6$ blocks, the details of which are given in Table \ref{tab:arch1}. The segmentation branch $f_\theta$ is identical to Deeplabv2 \cite{chen2017deeplab, french2019semi} with ResNet backbone \cite{he2016deep}. The subscript $\theta$ denotes the weights of all 6 blocks. The second branch $\hat{g}_{\hat\theta}$  (used only in Stage 2) shares the same structure and weights with $f_\theta$ up to the block \texttt{conv4}, and then has its own parameters on the last two blocks \texttt{conv6} and \texttt{ASPP\_2}. The block \texttt{conv6} has the same structure as \texttt{conv5}, and in order to reduce the training time, we use only half the number of channels. The weights of the $6$ blocks are denoted by $\hat{\theta}$.  Finally, the last branch $\tilde{g}_{\tilde{\theta}}$ (used only in Stage 3) shares the weights with $f_\theta$ up to the last but one block (\texttt{conv5}), and has its own classification module (\texttt{ASPP\_3}). Having many common blocks for the branches, the multi-task network has only slightly more parameters and floating point operations compared to the underlying segmentation network. 

\begin{table*}
\renewcommand\arraystretch{1.1}
\setlength{\tabcolsep}{3pt}
\begin{center}
\begin{tabular}{ccc}
\begin{tabular}{c|c}
\hline
\multicolumn{2}{c}{\cellcolor[HTML]{EFEFEF} $f_\theta$} \\
\hline
block & specification \\
\hline
\hline
\texttt{conv1} & \text{7$\times$7, stride 2} \\
\hline
\multirow{4}{*}{\texttt{conv2}} & \text{3$\times$3 max pooling, stride 2} \\
 &\blockb{256}{64}{3}  \\
&\\
& \\
\hline
\multirow{3}{*}{\texttt{conv3}} & 
\blockb{512}{128}{4} \\
&\\
& \\
\hline
\multirow{3}{*}{\texttt{conv4}} & 
\blockb{1024}{256}{23}  \\
&\\
& \\
\hline
\multirow{3}{*}{\texttt{conv5}} &
\blockb{2048}{512}{3}  \\
&\\
& \\
\hline
\texttt{ASPP\_1} & 3$\times$3, $C$, rate (6,12,18,24) \\
\end{tabular}
&
\begin{tabular}{c|c}
\hline
\multicolumn{2}{c}{\cellcolor[HTML]{EFEFEF} $\hat{g}_{\hat\theta}$} \\
\hline
block & specification \\
\hline
\texttt{conv1} & \text{7$\times$7, stride 2} \\
\hline
\multirow{4}{*}{\texttt{conv2}} & \text{3$\times$3 max pooling, stride 2} \\
 &\blockb{256}{64}{3}  \\
&\\
& \\
\hline
\multirow{3}{*}{\texttt{conv3}} & 
\blockb{512}{128}{4} \\
&\\
& \\
\hline
\multirow{3}{*}{\texttt{conv4}} & 
\blockb{1024}{256}{23}  \\
&\\
& \\
\hline
\multirow{3}{*}{\texttt{conv6}} &
\blockb{1024}{256}{3}  \\
&\\
& \\
\hline
\texttt{ASPP\_2} & 3$\times$3, $C$,  rate (6,12,18,24) \\
\end{tabular}
& 
\begin{tabular}{c|c}
\hline
\multicolumn{2}{c}{ \cellcolor[HTML]{EFEFEF} $\tilde{g}_{\tilde{\theta}}$} \\
\hline
block & specification \\
\hline
\texttt{conv1} & \text{7$\times$7, stride 2} \\
\hline
\multirow{4}{*}{\texttt{conv2}} & \text{3$\times$3 max pooling, stride 2} \\
 &\blockb{256}{64}{3}  \\
&\\
& \\
\hline
\multirow{3}{*}{\texttt{conv3}} & 
\blockb{512}{128}{4} \\
&\\
& \\
\hline
\multirow{3}{*}{\texttt{conv4}} & 
\blockb{1024}{256}{23}  \\
&\\
& \\
\hline
\multirow{3}{*}{\texttt{conv5}} &
\blockb{2048}{512}{3}  \\
&\\
& \\
\hline
\texttt{ASPP\_3} & 3$\times$3, $C$,  rate (6,12,18,24) \\
\end{tabular}
\\
(a) & (b) & (c) \\
\end{tabular}
\end{center}
\caption{The architecture of networks $f_\theta$, $\hat{g}_{\hat\theta}$ and $\tilde{g}_{\tilde\theta}$ for our experiments. Each network has $6$ blocks. The entries in the square blankets specify the size of convolutional kernels and the number of the output channels. The last block of the network is the Atrous Spatial Pyramid Pooling (ASPP) \cite{chen2017deeplab} with four different dilation rates. The number of final output channel is: $C=21$ for Pascal and $C=19$ for Cityscapes. The weights of \texttt{conv1},\texttt{conv2},..., \texttt{conv4} are shared by the three networks, while the block \texttt{conv5} is shared by  $f_\theta$ and $\tilde{g}_{\tilde\theta}$ }\label{tab:arch1}
\end{table*}

\section*{Acknowledgment}
RK and AAR acknowledge support from 
the EPSRC grant EP/T003553/1. AAR and CBS acknowledge support from the EPSRC Cambridge Mathematics of Information in Healthcare Hub EP/T017961/1. CBS also acknowledges support from
EU Horizon 2020 research and innovation programme NoMADS (Marie Skłodowska-Curie grant agreement No 777826),
the Philip Leverhulme Prize, 
the Royal Society Wolfson Fellowship,
the EPSRC grants EP/S026045/1,
EP/N014588/1,
the Wellcome Innovator Award RG98755, 
the Cantab Capital Institute for the Mathematics of Information and the Alan Turing Institute, and the Leverhulme Trust project Unveiling the invisible.

\ifCLASSOPTIONcaptionsoff
  \newpage
\fi

\bibliographystyle{IEEEtran}
\bibliography{egbib}

\end{document}